\documentclass[runningheads]{llncs}
\usepackage[T1]{fontenc}
\usepackage{array} % required for text wrapping in tables
\usepackage{graphicx}
\begin{document}
\title{AI-Powered Math Tutoring: Platform for Personalized and Adaptive Education}
\titlerunning{AI-Powered Math Tutoring}
% If the paper title is too long for the running head, you can set
% an abbreviated paper title here
%
\author{Jarosław A. Chudziak\orcidID{0000-0003-4534-8652}  \and\\ Adam Kostka\orcidID{0009-0004-8174-2109}}
\institute{Warsaw University of Technology, Poland
\\
\email{\{jaroslaw.chudziak, adam.kostka.stud\}@pw.edu.pl}}

% \author{Anonymous Author(s)}
% \institute{Paper submitted for anonymous review according to subguidelines}

%
\authorrunning{J. A. Chudziak \and A. Kostka}
% First names are abbreviated in the running head.
% If there are more than two authors, 'et al.' is used.
%
%
\maketitle              % typeset the header of the contribution
\begin{abstract}
The growing ubiquity of artificial intelligence (AI), in particular large language models (LLMs), has profoundly altered the way in which learners gain knowledge and interact with learning material, with many claiming that AI positively influences their learning achievements.
Despite this advancement, current AI tutoring systems face limitations associated with their reactive nature, often providing direct answers without encouraging deep reflection or incorporating structured pedagogical tools and strategies. This limitation is most apparent in the field of mathematics, in which AI tutoring systems remain underdeveloped.
This research addresses the question: How can AI tutoring systems move beyond providing reactive assistance to enable structured, individualized, and tool-assisted learning experiences?
We introduce a novel multi-agent AI tutoring platform that combines adaptive and personalized feedback, structured course generation, and textbook knowledge retrieval to enable modular, tool-assisted learning processes. This system allows students to learn new topics while identifying and targeting their weaknesses, revise for exams effectively, and practice on an unlimited number of personalized exercises.
This article contributes to the field of artificial intelligence in education by introducing a novel platform that brings together pedagogical agents and AI-driven components, augmenting the field with modular and effective systems for teaching mathematics.


\end{abstract}
\section{Introduction}

Mathematics education, vital for STEM fields as it provides foundational quantitative reasoning skills, faces challenges stemming from diverse learner needs, such as varying prior knowledge and learning paces, and the limitations of traditional settings \cite{van_Hoeve19082023,VISWANATHAN2022100057}, necessitating personalized approaches. Proficiency in mathematics is often seen as a gateway to innovation and critical thinking in these fields. Artificial intelligence (AI), especially Large Language Models (LLMs), shows potential for Intelligent Tutoring Systems (ITS) offering adaptive feedback and content generation \cite{wang2024largelanguagemodelseducation,llm_survey,Park_2024}.

However, many current LLM tutors provide direct answers, often overlooking the crucial cognitive steps necessary for effective learning and hindering deep understanding \cite{macina-etal-2023-opportunities}, lack robust integration with course materials \cite{supporting_teaching}, and offer limited dynamic adaptation \cite{Park_2024}. A gap exists for AI tutors that foster deep learning through guided discovery using structured knowledge \cite{macina-etal-2023-mathdial,supporting_teaching}. Our research addresses: \textit{How can AI tutoring systems move beyond reactive assistance to enable structured, individualized, and tool-assisted learning experiences?}

We introduce a multi-agent AI tutoring environment combining adaptive Socratic agents, dual-memory personalization, GraphRAG textbook retrieval, and Directed Acyclic Graph (DAG)-based course planning. Our approach prioritizes deep understanding and independent problem-solving over direct answers. Contributions include: a novel multi-agent architecture integrating these components; a benchmark-validated, pedagogically-driven conversational agent; and a modular, open-source infrastructure\footnote{\url{https://github.com/feilaz/AI\_Powered\_Math\_Tutoring}}.
%%%%%%%%%%%%%%%%%%%%%%%%%%%%%%%%%%%%%%%%%%%%%%%%%%%%%%%%%%

\section{Related Work}

Examples of early ITS include AutoTutor \cite{AutoTutor} and ASSISTments \cite{ASSISTments}. They were promising but held limitations. In contrast, LLMs broadened horizons of applications \cite{Mishra_Sutanto_Rossanti_Pant_Ashraf_Raut_Uwabareze_Oluwatomiwa_Zeeshan_2024}, but their characteristic of always giving answers directly goes against the goals of critical thinking \cite{10.1145/3627673.3679665,Wang_Xu_Zhang_Hao_Jie_2024}. Many benchmarks show the challenge in building guiding LLM tutors, MathDial \cite{macina-etal-2023-mathdial,macina-etal-2023-opportunities} being just one of them.

\begin{figure}[t]
\centering
% Use placeholder or actual path to image file
\includegraphics[width=0.9\linewidth]{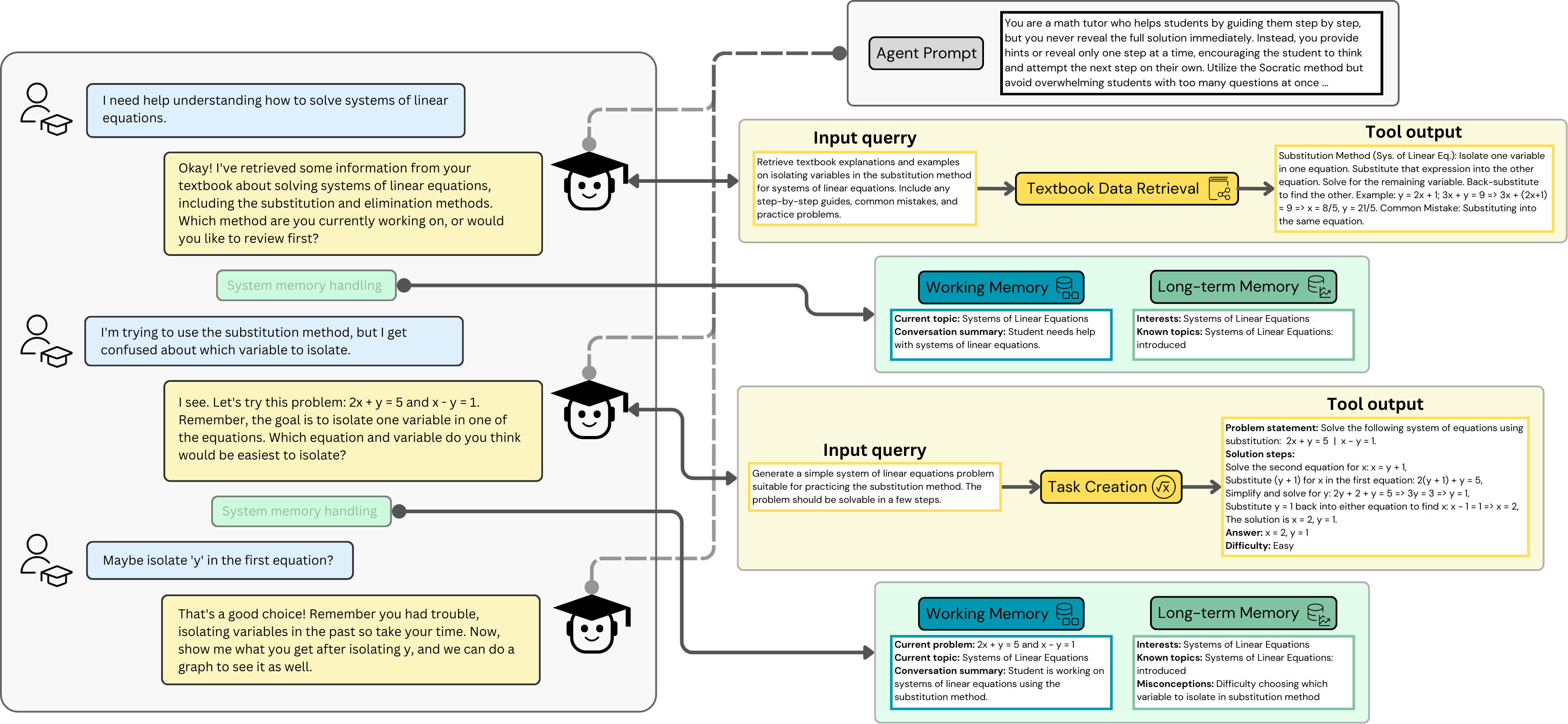}
\vspace{-2mm} % Optional adjustment
\caption{Guided tutoring interaction showing Socratic questioning and personalized support components.}
\label{fig:convo1}
\vspace{-3mm} % Optional adjustment
\end{figure}

Multi-agent systems (MAS) are a promising approach to complex tutoring \cite{VISWANATHAN2022100057,synergymas}, leveraging coordinated agents for specialized tasks, similar to applications seen in other complex domains like financial analysis \cite{paclic_elliottagents}. Even with the advances in mathematical reasoning \cite{Trinh_Wu_Le_He_Luong_2024,guan2025rstarmathsmallllmsmaster}, many MAS lack the necessary fine-grained knowledge integration and deep personalization required to tailor guidance precisely to evolving student understanding, which underpins our adaptive tutoring approach. Previous research on ITS architectures \cite{sonkar-etal-2023-class} and RAG assistants \cite{jill_watson} has contributed to our work.

Retrieval-Augmented Generation (RAG) is crucial for knowledge retrieval, grounding LLMs knowledge \cite{supporting_teaching}. We use GraphRAG \cite{edge2024from} based on the argument that its graph structure better represents educational content relations than normal vector RAG for contextual tutoring \cite{han2025ragvsgraphragsystematic}.

Though our work draws on areas, such as ITS design principles \cite{sonkar-etal-2023-class} and dialogue systems \cite{Park_2024}, it does so in a unique way by integrating these within a comprehensive multi-agent framework focusing on guided mathematical learning, using GraphRAG and dual-memory personalization.

%%%%%%%%%%%%%%%%%%%%%%%%%%%%%%%%%%%%%%%%%%%%%%%%%%%%%%%%%%

\begin{figure}[t]
\centering
% Use placeholder or actual path to image file
\includegraphics[width=1\linewidth]{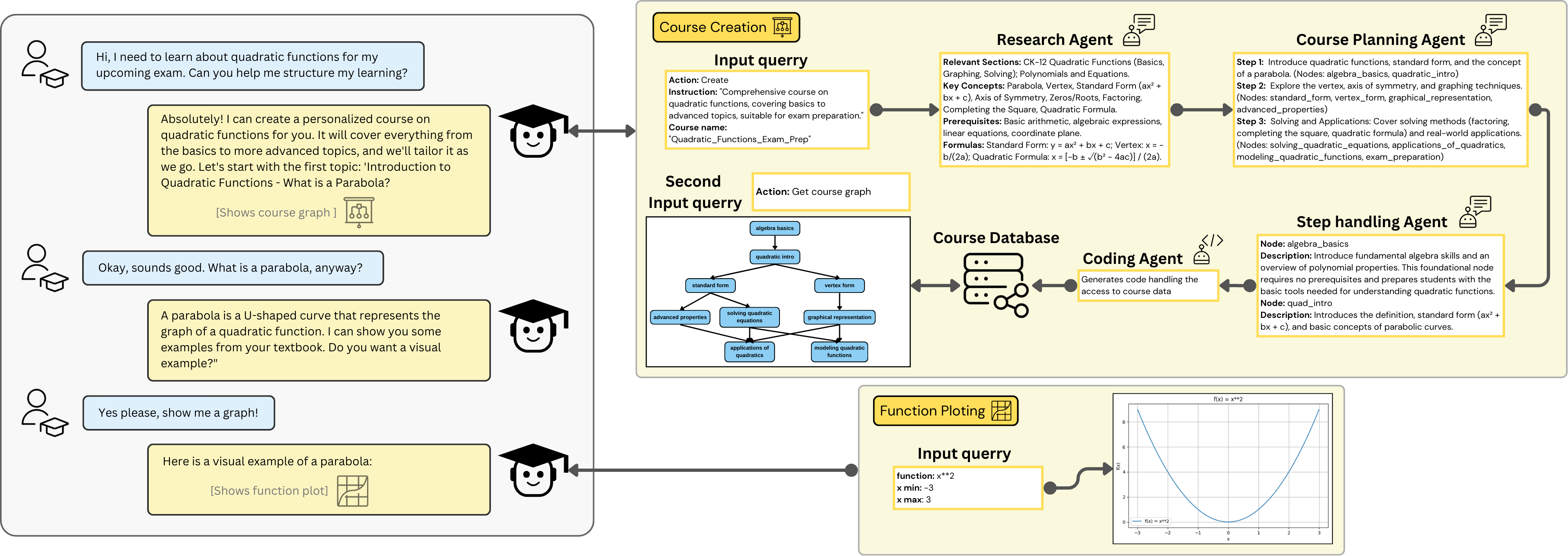}
\vspace{-2mm} % Optional adjustment
\caption{Course creation using a DAG structure for structured learning and targeted practice.}
\label{fig:convo2}
\vspace{-3mm} % Optional adjustment
\end{figure}

\section{System in Action: User Stories}

This section provides exemplary system functionalities through user stories. It presents the components of Intelligent Textbook Retrieval (GraphRAG), the Tutoring Agent, Personalization (Memory), the Course module, and Task Creation.

\subsection{Guided and Personalized Tutoring with AI}
\label{subsec:guided_tutoring}

Guided learning is the principle of the system, contrary to giving direct answers. Instead of solving an equation immediately, the Tutoring Agent uses Socratic questioning (Figure \ref{fig:convo1}), promoting self-explanation and metacognitive skills, dynamically adjusting support based on progress tracked in memory. The agent retrieves textbook data ("Textbook Data Retrieval") to ground its responses and generates practice tasks ('Task Creation') as needed, fostering a deeper understanding rather than rote learning.

Personalization uses Long-Term Memory (LTM) for stable traits (e.g., topic mastery, common misconceptions, learning style) and Working Memory (WM) for the current session context. For instance, if LTM knows that a student usually has difficulty understanding negative sign distribution, the system might proactively offer targeted hints, scaffold the problem differently, or provide corrective feedback when the error occurs. On the same principle, if LTM notes that a student prefers visual explanations, aids such as function plotting will be used, creating a responsive and individualized learning experience.

\subsection{Course Creation for Exam Revision}
\label{subsec:course_creation}

For any structured learning, such as exam preparation, the Course component will generate an individual Directed Acyclic Graph (DAG) concerning relevant topics (as shown in Figure \ref{fig:convo2}). Such a DAG shows prerequisite knowledge, suggests optimal learning paths based on textbook knowledge and dependency analysis, and enables students to monitor their learning progress effectively. A student preparing for an examination may request a revision course, and the system automatically arranges a structured plan individualized to their specific needs.

%%%%%%%%%%%%%%%%%%%%%%%%%%%%%%%%%%%%%%%%%%%%%%%%%%%%%%%%%%

\begin{figure}[t]
    \centering
    % Use placeholder or actual path to image file
    \includegraphics[width=0.9\linewidth]{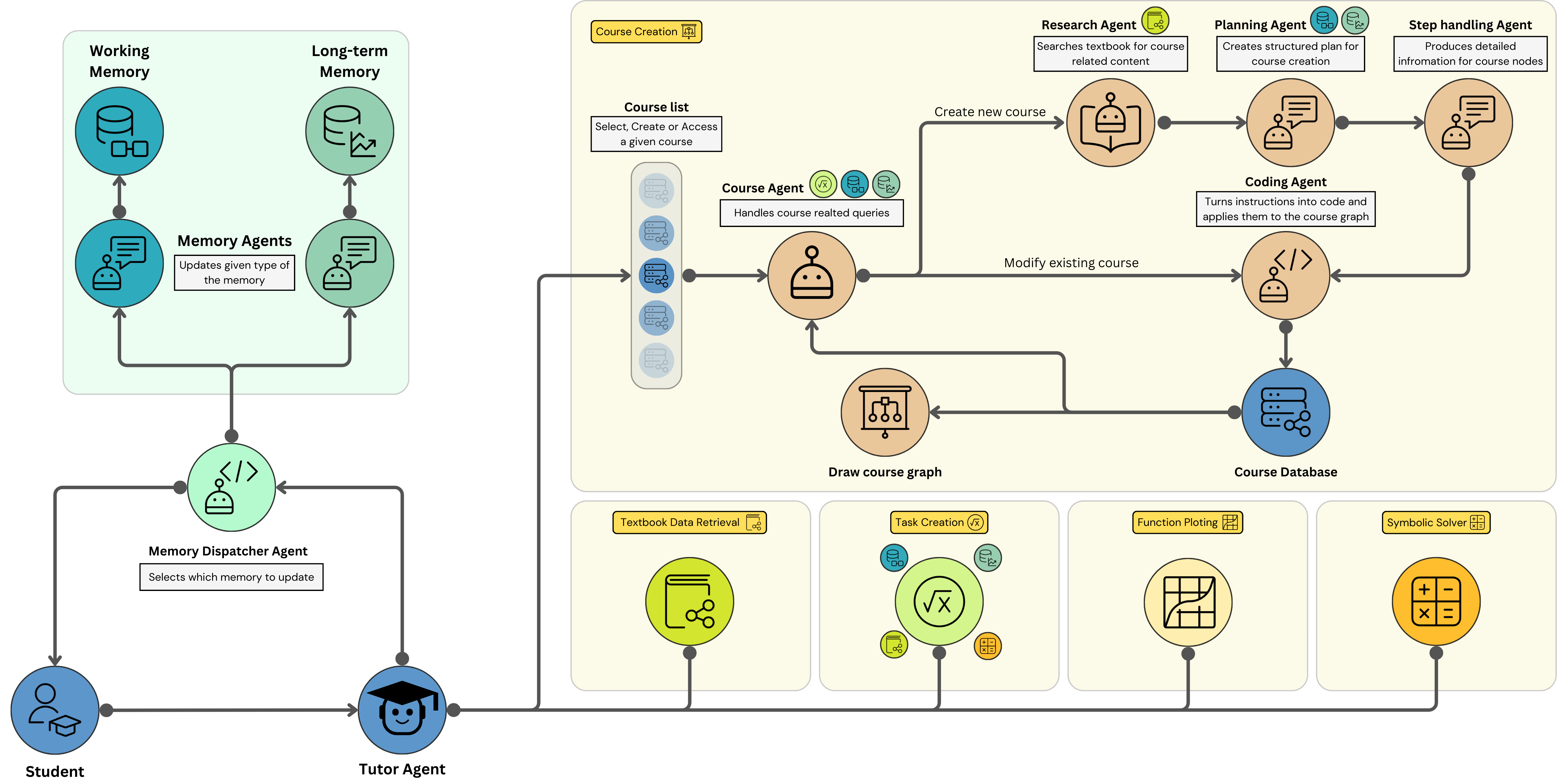}
    \vspace{-2mm} % Optional adjustment
    \caption{System architecture: Information flow between Student, Tutor Agent, Memory, Specialist Agents (Research, Planning, etc.), and Tools (Retrieval, Task Creation, Solver, etc.).}
    \label{fig:use_cases}
    \vspace{-3mm} % Optional adjustment
\end{figure}

\section{System Architecture and Functional Overview}

A multi-agent architecture implemented on LangGraph allows for adaptive learning through interconnected components, as shown in Figure \ref{fig:use_cases}. The main interaction loop revolves around the Tutor Agent (GPT-4o), chosen for its conversational capabilities, proficiency in tool use, and low latency. This agent interprets the student input and orchestrates the other components via a ReAct-style framework \cite{yao2023reactsynergizingreasoningacting}. A Memory Dispatcher oversees conversations and controls the appropriate memory modules for personalization.

\subsection{System Components}

The dual-memory framework that supports adaptive learning distinguishes between long-term and working memory. Long-term memory (LTM) stores persistent, student-specific data (e.g., prior knowledge, misconceptions, learning preferences, goals), while working memory (WM) maintains information regarding the context of the current learning session (e.g., topic, problem state, recent interactions). These memory modules inform the Tutor Agent and content generation components. To ensure access to structured educational content, the system implements Retrieval-Augmented Generation via the GraphRAG framework \cite{edge2024from}, where textbook material is represented as a knowledge graph to provide contextually relevant information.

The system is responsible for generating both individual exercises and structured courses. A specific Task Creation module harnesses the o3-mini(high) model (selected for its strong mathematical reasoning, validated in Section \ref{sec:evaluation}) to generate personalized practice exercises based on topic, difficulty, and optionally using GraphRAG data. To create courses, a pipeline of agents (Research -> Planning -> Step Handling -> Coding agents) accesses memory and GraphRAG data to construct a structured DAG guiding the learning process and stored in a database \cite{10764913}.
Additionally, the system includes auxiliary tools, such as the Symbolic Solver based on SymPy, a Function Plotter based on Matplotlib, and a Course Graph Drawer.

%%%%%%%%%%%%%%%%%%%%%%%%%%%%%%%%%%%%%%%%%%%%%%%%%%%%%%%%%%

\begin{figure}[t]
    \centering
    % Use placeholder or actual path to image file
    \includegraphics[width=1\textwidth]{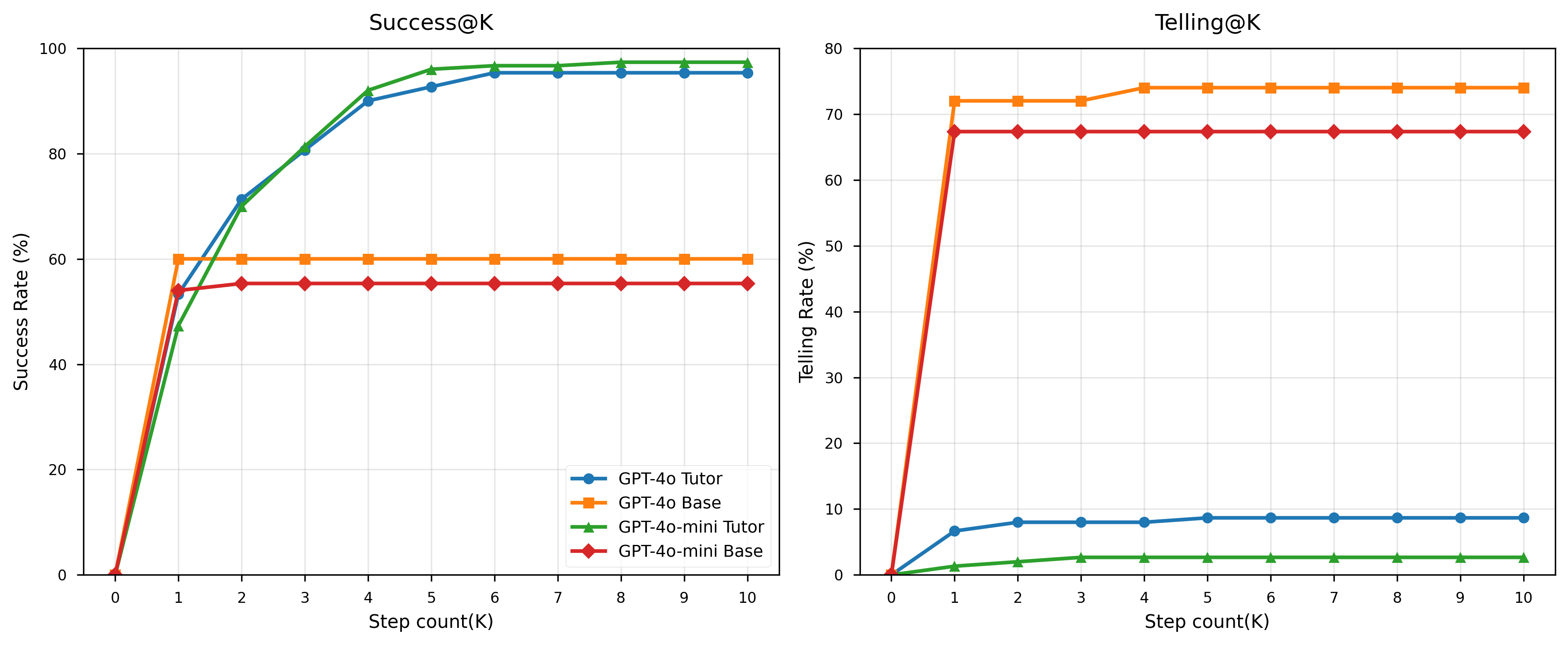}
    \vspace{1em} % Space between plot and table
    \caption{Guided Tutoring Benchmark (Mathdial): Tutor Prompt yields higher Success@K and lower Telling@K than Base Prompt, validating its guidance effectiveness.}
    \label{fig:evaluation}
\end{figure}

\section{Evaluation}
\label{sec:evaluation}

Evaluation of major design aspects using the MathDial dataset \cite{macina-etal-2023-mathdial} served both to validate our pedagogical approach (guided tutoring) and to inform the selection of models relative to specific system components.

We assessed the Tutor Agent's ability to lead learning without prematurely revealing the answers (low 'Telling@N') while still driving the student to a successful solution ('Success@N'). We compared GPT-4o and GPT-4o-mini using our pedagogically informed 'Tutor Prompt' (which emphasized Socratic questioning and scaffolding) against MathDial's basic 'Base Prompt', thereby simulating a tutor-student interaction without external tools.

Results (shown in Figure \ref{fig:evaluation}) displayed that the 'Tutor Prompt' significantly outperformed the 'Base Prompt' for each model, achieving superior Success@N and far lower Telling@N rates over interaction lengths (K). This validates our prompting strategy's effectiveness in promoting guided tutoring compared to direct question answering in simulated dialogues.

To choose the model fitting for mathematical intensive Task Creation and Course Planning, we tested the problem-solving accuracy of various LLMs(with access to the SymPy tool) on the MathDial problem set. Model o3-mini(high) and Claude 3.5 Sonnet obtained the highest accuracy, of 90.00\%, followed by Gemini 2.0 Flash (88.67\%) with GPT-4o (78.67\%) and GPT-4o-mini (77.33\%) obtaining the lowest scores. Consequently, o3-mini(high) was selected for the Task Creation component, given its top performance and step-decomposition ability.

%%%%%%%%%%%%%%%%%%%%%%%%%%%%%%%%%%%%%%%%%%%%%%%%%%%%%%%%%%

\section{Discussion and Future Work}

Our assessment confirms key design choices, particularly the effectiveness of guided tutoring prompts. However, a significant limitation noted by reviewers is the lack of evaluation with real students in learning environments. Current personalization operates on a basic set of attributes, while the pedagogical effectiveness of courses generated through GraphRAG for students is yet to be evaluated empirically. System performance remains fundamentally dependent on underlying LLM capabilities and possible biases.

Future work will focus on real-world user studies in order to assess learning gains, the user experience, and the practical utility of generated courses. We aim to improve personalization through advanced student modeling (e.g., adding affective state detection) and integrate mechanisms such as spaced repetition. Technical improvements will also explore the potential of different LLMs and RAG architectures \cite{liang2024kagboostingllmsprofessional}. Future work will also research adaptability to other STEM domains, accessibility \cite{Liu2024ScaffoldingLL}, and explainability.

%%%%%%%%%%%%%%%%%%%%%%%%%%%%%%%%%%%%%%%%%%%%%%%%%%%%%%%%%%

\section{Conclusion}

We presented a multi-agent AI math tutor integrating adaptive tutoring, personalization, GraphRAG, and course planning to reach deep understanding through guided discovery. Contributions include the novel MAS architecture, validated agent design, and the open-source framework. Although user validation is required, it demonstrates the potential for combining AI techniques and pedagogical methods for effective personalized STEM learning. Successfully addressing these challenges could significantly enhance STEM education accessibility and outcomes. We plan to refine the system by incorporating future real-world feedback.

\subsubsection{\ackname}
We would like to acknowledge that the work reported in this paper has been supported in part by the Polish National Science Centre, Poland (Chist-Era IV) under grant 2022/04/Y/ST6/00001

%
% ---- Bibliography ----
%
% BibTeX users should specify bibliography style 'splncs04'.
% References will then be sorted and formatted in the correct style.
%
\bibliographystyle{splncs04}
% Ensure the .bib file is heavily curated to fit within 2 pages.
\bibliography{mybibliography} % Assumes 'mybibliography.bib' is curated

\end{document}